\documentclass{article}
\usepackage[preprint]{neurips_2025}
\usepackage[utf8]{inputenc} 
\usepackage[T1]{fontenc}    
\usepackage{url}            
\usepackage{booktabs}       
\usepackage{amsfonts}       
\usepackage{nicefrac}       
\usepackage{microtype}      
\usepackage{soul}
\usepackage{graphicx}
\usepackage{amsmath}
\usepackage{outlines}
\usepackage{xurl}

\usepackage[colorlinks=true,
citecolor=blue,
filecolor=black,
linkcolor=blue,
urlcolor=blue]{hyperref}

\usepackage{amsmath,amsfonts,bm}

\usepackage{xcolor}
\usepackage{colortbl}

\def\eqref#1{equation~\ref{#1}}

\DeclareMathAlphabet{\mathsfit}{\encodingdefault}{\sfdefault}{m}{sl}
\SetMathAlphabet{\mathsfit}{bold}{\encodingdefault}{\sfdefault}{bx}{n}

\usepackage{multirow}
\usepackage{paralist}

\usepackage{multicol}
\usepackage{diagbox}

\usepackage[ruled,noend]{algorithm2e}

\SetCommentSty{mycommfont}

\usepackage{here}

\usepackage{amsmath,amssymb,amsfonts,amsbsy,amsfonts,latexsym}
\usepackage{makecell}
\usepackage{xcolor}
\usepackage{colortbl}

\usepackage{tabularx,colortbl,xcolor}
\usepackage[normalem]{ulem}
\useunder{\uline}{\ul}{}

\usepackage{enumitem}

\usepackage{xparse}

\SetKwInput{KwInput}{Input}
\SetKwInput{KwRequire}{Require}

\usepackage{longtable}

\usepackage{bigdelim}

\NewDocumentCommand{\var}{O{s} m O{}}{%
  \ensuremath{#1_{#2}^{#3}}
}
\usepackage{siunitx}

\newcommand{\commentout}[1]{}

\definecolor{light-gray}{gray}{0.80}

\newcommand\appref{Appendix~\ref}

\newcommand\fref{Figure~\ref}
\newcommand\tref{Table~\ref}
\newcommand\sref{Section~\ref}

\usepackage{amsthm}

\usepackage{amsthm}

\newtheorem{mytheorem}{Theorem}
\newtheorem{assumption}[mytheorem]{Assumption}
\newtheorem{lemma}[mytheorem]{Lemma}
\newtheorem{remk}[mytheorem]{Remark}

\def\huggingface{\raisebox{-1.5pt}{\includegraphics[height=1.05em]{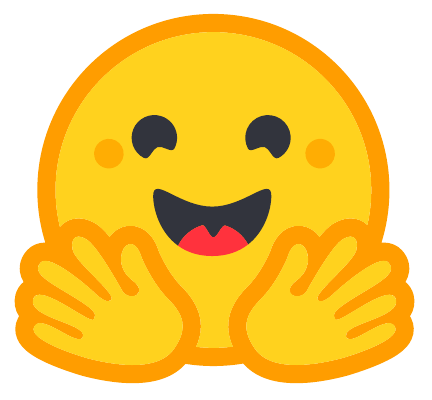}}}
\def\github{\raisebox{-1.5pt}{\includegraphics[height=1.05em]{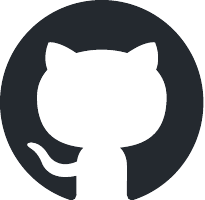}}}

\newcommand{\hflink}{https://huggingface.co/Snowflake/Arctic-Text2SQL-R1-7B}
\newcommand{\ghlink}{https://github.com/snowflakedb/ArcticTraining/}

\usepackage{subfig}

\usepackage{pifont}
\usepackage{footmisc}

\newcommand{\cmark}{\ding{51}}%
\newcommand{\xmark}{\ding{55}}%

\usepackage[most]{tcolorbox}
\usepackage{xcolor}  
\newtcolorbox{prompt}[1]{
  enhanced,
  colback=blue!5!gray!10,
  colframe=blue!50!black,
  arc=2mm,
  boxrule=1pt,
  title=#1,
  fonttitle=\bfseries\color{white},
  coltitle=green!50!black,
  breakable=false,
  width=\textwidth
}

\usepackage{geometry}
\usepackage{inconsolata}
\newtcolorbox{promptbox}[1][]{
  colback=gray!5,
  colframe=black!50,
  boxrule=0.8pt,
  arc=2mm,
  fontupper={\ttfamily\small},
  enhanced,
  breakable,
  left=5pt,
  right=5pt,
  top=5pt,
  bottom=5pt,
  title=#1
}

\newcommand{\OURS}{Arctic-Text2SQL-R1\xspace}
\newcommand{\oursb}{Arctic-Text2SQL-R1-7B\xspace}
\newcommand{\ourfb}{Arctic-Text2SQL-R1-14B\xspace}
\newcommand{\ourttb}{Arctic-Text2SQL-R1-32B\xspace}

\newcommand{\tts}{Text2SQL\xspace}
\newcommand{\grpo}{GRPO\xspace}
\newcommand{\ppo}{PPO\xspace}
\newcommand{\bird}{BIRD\xspace}
\newcommand{\spider}{SPIDER\xspace}
\newcommand{\spidertwo}{Spider2.0\xspace}
\newcommand{\spiderdk}{Spider-DK\xspace}
\newcommand{\gretel}{Gretel-Synth\xspace}
\newcommand{\ehrsql}{EHRSQL\xspace}
\newcommand{\science}{ScienceBenchmark\xspace}
\newcommand{\qwq}{QwQ-32B\xspace}
\newcommand{\omnisql}{OmniSQL\xspace}
\newcommand{\omnisqlsb}{OmniSQL-7B\xspace}

\newcommand{\qwensb}{Qwen2.5-Coder-7B-Inst\xspace}
\newcommand{\qwenfb}{Qwen2.5-Coder-14B-Inst\xspace}
\newcommand{\qwenttb}{Qwen2.5-Coder-32B-Inst\xspace}

\usepackage{chngcntr}
\usepackage{adjustbox}

\newcommand\atsign{@}

\makeatletter
\newcommand\thanksnomark[1]{%
  \begingroup
  \renewcommand\@makefnmark{}
  \thanks{#1}%
  \endgroup
}
\makeatother

\begin{document}

\title{\OURS: Simple Rewards, Strong Reasoning in Text-to-SQL}

\author{Zhewei Yao$^{1,*}$\thanksnomark{Project lead. Corresponding to zhewei.yao{\atsign}snowflake.com.}, Guoheng Sun$^2$, Lukasz Borchmann$^1$, Gaurav Nuti$^1$, Zheyu Shen$^2$, \\
\bf{Minghang Deng$^3$, Bohan Zhai$^1$, Hao Zhang$^{1,3}$, Ang Li$^2$, Yuxiong He$^1$}\\
$^1$Snowflake AI Research, $^2$University of Maryland, Colleage Park, \\$^3$University of California, San Diego}

\maketitle

\begin{center}
\vspace{-0.5cm}
\begin{tabular}{rl}
\huggingface & \url{\hflink}\\
\github & \url{\ghlink}\\
\end{tabular}
\vspace{0.5cm}
\end{center}

\begin{abstract}

Translating natural language into SQL (\tts) is a longstanding challenge at the intersection of natural language understanding and structured data access. While large language models (LLMs) have significantly improved fluency in SQL generation, producing correct and executable SQL—particularly for complex queries—remains a bottleneck. We present \textbf{\OURS}, a reinforcement learning (RL) framework and model family designed to generate accurate, executable SQL using a lightweight reward signal based solely on execution correctness. Our approach avoids brittle intermediate supervision and complex reward shaping, promoting stable training and alignment with the end task. Combined with carefully curated data, strong supervised initialization, and effective training practices, \OURS achieves state-of-the-art execution accuracy across six diverse \tts benchmarks, including the top position on the BIRD leaderboard. Notably, our 7B model outperforms prior 70B-class systems, highlighting the framework’s scalability and efficiency. We further demonstrate inference-time robustness through simple extensions like value retrieval and majority voting. Extensive experiments and ablation studies offer both positive and negative insights, providing practical guidance for future \tts research.

\end{abstract}
\section{Introduction}
\label{sec:intro}

Translating natural language questions into SQL queries—commonly known as \textbf{\tts}—is a core challenge in natural language understanding and human-computer interaction. A robust solution would enable non-technical users to query structured databases using natural language, thereby democratizing access to data analytics.

Recent advances in large language models (LLMs) have significantly improved the fluency and surface-level coverage of \tts systems~\cite{ma2025sql,pourreza2025reasoning,zhai2025excot,papicchio2025think2sqlreinforcellmreasoning}. However, generating \emph{correct and executable} SQL—especially for complex queries involving multi-table joins, nested logic, and nuanced schema understanding—remains difficult. Most existing approaches rely on supervised fine-tuning over (question, SQL) pairs~\cite{omnisql,gao2024xiyan}, which often fail to promote the intermediate reasoning steps essential for reliable and generalizable SQL generation.

\begin{table}[t]
\begin{minipage}[t]{.405\linewidth}
\caption{\bird \href{https://bird-bench.github.io/}{Leaderboard} Single-Model Leaderboard.\protect\footnotemark}
\centering
\label{tab:bird_leaderboard}
\begin{adjustbox}{width=0.99\linewidth}
\centering
\begin{tabular}{lrr }
        \toprule
        Model (Rank)& Dev & Test \\
        \midrule
        \OURS-32B (Top-1)&70.5 &71.8\\
        Infly-RL-SQL-32B (Top-2) & 70.1 & 70.6 \\
        \OURS-14B (Top-3)&70.1 &70.0\\
        XiYan-32B~\cite{gao2024xiyan} (Top-4)&67.0 & 69.0\\
        Arctic-ExCoT-70B~\cite{zhai2025excot} (Top-5) & 68.5 & 68.5\\
        \OURS-7B (Top-6)&68.9 &68.5\\
        Command-A-111B~\cite{cohere2025commanda} (Top-8) &63.5 &65.7\\
        \bottomrule
\end{tabular}
\end{adjustbox}
\end{minipage}\hspace{1em}\hfill
\begin{minipage}[t]{.585\linewidth}
\caption{The reward design comparison of \OURS to existing reasoning models.}
\centering
\label{tab:techniques_comparison}
\begin{adjustbox}{width=0.99\linewidth}
\centering
\setlength{\tabcolsep}{3pt}
\renewcommand{\arraystretch}{1.11}
\begin{tabular}{ll }
        \toprule
        Framework & Optimization (Reward) \\
        \midrule
        Reasoning-SQL~\cite{pourreza2025reasoning}
        & GRPO {\footnotesize(EX, syntax, n-gram, LLM, schema, format)}\\
        SQL-R1~\cite{ma2025sql}
        & GRPO {\footnotesize(EX, length, syntax, format)}\\
        Think2SQL~\cite{papicchio2025think2sqlreinforcellmreasoning}
        & GRPO {\footnotesize(precision, recall, cardinality, 2$\times$format)} \\
        ExCoT \cite{zhai2025excot}& DPO ~~{\footnotesize(EX)} \\
        \midrule
        \OURS
        & GRPO {\footnotesize(EX, syntax)}
        \\
        \bottomrule
\end{tabular}
\end{adjustbox}
\end{minipage}
\end{table}

\looseness=-1 We introduce \textbf{\OURS}, a reinforcement learning (RL) framework and model family for generating high-quality executable SQL from natural language. 
It uses a lightweight reward based solely on execution correctness, avoiding brittle partial rewards while promoting stable training and alignment with the end goal. 
Combined with best practices in data filtering, synthetic generation, and model-based selection, 
\OURS demonstrates that high-quality data, strong supervised initialization, and a simple reward signal together yield accurate, scalable, and robust \tts models.

\textbf{Our main contributions are:}

\begin{itemize}[noitemsep, nolistsep, labelindent=0pt, leftmargin=20pt]
    \item \textbf{Simple and Scalable RL for \tts:} We develop a lightweight RL formulation using execution-only rewards that generalizes well across model sizes and benchmarks, enabling stable training and strong performance without complex reward design.
    
    \item \textbf{State-of-the-Art Performance Across Benchmarks:} \OURS achieves \#1 execution accuracy on the BIRD leaderboard, with our 32B model reaching 71.83\% and the 14B model surpassing the 70\% milestone—a first among open models. Across six diverse \tts benchmarks, \OURS consistently outperforms both SQL-specialized and general-purpose LLMs, including GPT-4o~\cite{openai2023gpt4} and DeepSeek-V3~\cite{liu2024deepseek}. Notably, our 7B model outperforms prior 70B-class models~\cite{zhai2025excot}, demonstrating scalability and efficiency.
    
    \item \textbf{Data and Training Strategies That Matter:} We present best practices for data filtering, synthetic data generation, and model-based selection. Additionally, we demonstrate the value of \emph{online RL training}, which leverages strong supervised checkpoints and well-designed prompts to further improve performance. Together, these practices—when combined with high-quality data and a simple execution-driven reward—yield accurate, scalable, and robust \tts models.
    
    \item \textbf{Broad and Rigorous Evaluation:} We evaluate \OURS across six \tts datasets, covering a range of schema complexity and query difficulty. This provides strong evidence of generalization and guards against overfitting to any single dataset or evaluation format.
    
    \item \textbf{Inference-Time Robustness and Extensibility:} \OURS supports simple inference-time techniques—such as value retrieval and majority voting—that further improve accuracy with minimal system overhead, demonstrating its practicality for real-world deployment.
    
    \item \textbf{Empirical Insights for the Community:} We report both successful and negative findings, offering transparent and actionable insights to guide future RL-based \tts research/development.
\end{itemize}

\footnotetext{There is no clear instruction from \bird leaderboard to state that if greedy/single-generation decoding is a hard requirement. Therefore, we have less information about the settings of other models.}

\section{Related Work}\label{sec:related_work}  
   
The research presented in \OURS draws inspiration from and aims to advance two key areas of investigation: the development of robust \tts systems and the refinement of reinforcement learning techniques for sophisticated language model reasoning.

\textbf{Reinforcement Learning for LLM Reasoning.}  
\looseness=-1 Recent research has demonstrated the potential of RL techniques to significantly enhance the reasoning capabilities of large language models (LLMs)~\cite{openai2025competitiveprogramminglargereasoning,Lambert2024TLU3P,openai2024openaio1card,guo-etal-2019-towards}. By providing explicit rewards for logical correctness and step-wise reasoning, RL enables models to autonomously discover effective problem-solving strategies, often mirroring structured human reasoning~\cite{xu2025largereasoningmodelssurvey,wang2025enhancingcodellmsreinforcement,yang2025codethinkthinkcode}. Applications span mathematical problem solving (where RL fine-tuning on step-by-step correctness or final answer accuracy yields substantial improvements~\cite{shao2024deepseekmath,openai2024openaio1card}) and code generation, where preference optimization and RL from human feedback have led to greater code validity and efficiency~\cite{wang2025enhancingcodellmsreinforcement,yang2025codethinkthinkcode}.  
   
Most prior methods are built atop policy gradient algorithms such as Proximal Policy Optimization (PPO)~\cite{schulman2017proximalpolicyoptimizationalgorithms} or, more recently, Group Relative Policy Optimization (GRPO)~\cite{guo2025deepseek,Yang2024Qwen25MathTR,Dang2025ReinforcementLF}, which compares groups of generated responses rather than evaluating them in isolation. This approach is particularly powerful for reasoning tasks with multiple plausible solutions, enabling the model to build a deeper understanding of what constitutes high-quality reasoning~\cite{kumar2025llmposttrainingdeepdive}. For example, GRPO has enabled models like DeepSeek-R1-Zero to develop complex reasoning skills such as multi-step chain-of-thought and self-verification simply by being rewarded for correct final answers~\cite{guo2025deepseek}.
   
The effectiveness of these RL frameworks often hinges on carefully designed preference datasets and reward models that accurately reflect the subtleties of logical reasoning. Our work draws on these insights, by adapting GRPO for the \tts domain and focusing the reward signal on final executable correctness, encouraging the model to reason through the full compositional structure of the query.  
   
\textbf{Text-to-SQL.}
The \tts task has seen remarkable progress over the years~\cite{zhu2024largelanguagemodelenhanced}. Early systems were built on rule-based approaches and handcrafted grammars, but proved brittle when faced with linguistic ambiguity or complex schema variation~\cite{androutsopoulos1995naturallanguageinterfacesdatabases,10.14778/2735461.2735468}. The advent of neural sequence-to-sequence models helped automate parts of the semantic parsing process, though such models often required elaborate schema encoders and strong supervision~\cite{guo-etal-2019-towards,wang2021ratsqlrelationawareschemaencoding}.  
   
\looseness=-1 More recently, the generalization power of LLMs has revolutionized the field, as demonstrated by numerous works leveraging pre-trained LLMs for template-free SQL generation~\cite{singh2025surveylargelanguagemodelbased,gao2023texttosqlempoweredlargelanguage,li2024can}. LLMs often exhibit basic SQL competence on ``out-of-the-box'' evaluation, but their performance on complex, multi-table queries remains limited unless explicit compositional reasoning is encouraged~\cite{wei2023chainofthoughtpromptingelicitsreasoning,shao2024deepseekmath,guo2025deepseek,openai2024openaio1card}. This has motivated the development of techniques such as Chain-of-Thought prompting~\cite{tai2023exploringchainofthoughtstyleprompting}, query decomposition~\cite{eyal-etal-2023-semantic}, optimization~\cite{zhai2025excot}, as well as dynamic schema linking
and execution-based feedback~\cite{hong2024knowledgetosqlenhancingsqlgeneration,deng2025reforcetexttosqlagentselfrefinement}.  
   
Within this landscape, there is growing interest in RL-based approaches for \tts, often building on structured, multi-component reward functions that aggregate execution feedback, syntactic validity, partial string overlap, schema conformance, and more~\cite{pourreza2025reasoning,ma2025sql,papicchio2025think2sqlreinforcellmreasoning}. However, such complex reward engineering risks encouraging superficial reward hacking, as observed in RL for semantic parsing~\cite{skalse2022defining}. In contrast, our work advocates for a simpler, execution-centric reward design, inspired by earlier RL-based semantic parsing approaches such as Seq2SQL~\cite{zhong2017seq2sqlgeneratingstructuredqueries}, and demonstrates that this minimal reward scheme can be both more stable and more effective in practice.  
   
Compared to most recent reasoning works~\cite{pourreza2025reasoning,ma2025sql,he2025starsqlselftaughtreasonertexttosql}, or pairwise preference optimization~\cite{zhai2025excot}, our method is unique in unifying a streamlined data filtering pipeline, group-based relative policy optimization, and a strictly execution-based reward (Table~\ref{tab:techniques_comparison}).

\begin{table}[t]
\caption{
The datasets and their volume used in the paper. For \bird-training, spider-training/dev, their original sample sizes are 9428, 8659/1034.
}\centering
\label{tab:dataset}
\begin{adjustbox}{width=0.99\linewidth}
\centering
\begin{tabular}{lcccccccccccccc }
\toprule
  &\multicolumn{4}{c}{Used for Training} &\multicolumn{6}{c}{Used for Evaluation} \\
  \cmidrule(r){2-5} \cmidrule(l){6-11}
\multirow{2}{*}{Dataset} &\bird & \spider- & \spider- &\gretel & \bird- & \spider- &Spider2.0- &Spider- & EHR & Science \\ 
&training & training & dev &Filtered & dev & test &SQLite & DK & SQL & Benchmark  \\
\midrule
Size &8 017&6 972&985&11 811 &1534 &2 147 &135 &535 &1 008 &299\\
\bottomrule
\end{tabular}
\end{adjustbox}
\end{table}

\section{Methodology}\label{sec:methodology}  
   
Building on insights from prior reinforcement learning (RL) work for language model reasoning, we design \OURS to pursue robust \tts generation via a simple, execution-grounded RL framework. 

\textbf{Overview of RL Approach.} We adopt GRPO~\cite{shao2024deepseekmath} as it has demonstrated superior efficiency and effectiveness on structured reasoning tasks, making it an ideal match for \tts, where compositional accuracy and performance stability are crucial.

Formally, let $\pi_\theta$ denote our policy model parameterized by $\theta$. For each input text question $Q$ with associated database schema, the model generates $N$ candidate SQL queries (aka rollouts), $\{o_{Q,1}, ..., o_{Q,N}\}$. Each generated query is then evaluated to provide an explicit reward signal as described in the later of the section. 
These per-group rollouts allow us to compute relative advantages, stabilizing learning and promoting robust policy improvement. 
   
The GRPO objective is as follows:  
\begin{equation*}  
\mathcal{J}_{\text{GRPO}}(\theta) = \mathbb{E}\bigg[  
  \frac{1}{N} \sum_{i=1}^N \min\left(r_i A_i, \; \mathrm{clip}(r_i, 1-\epsilon, 1+\epsilon)\right)  
\bigg]  
- \beta \mathrm{D_{KL}}(\pi_\theta \| \pi_{ref}),  
\end{equation*}  
where $r_i = \frac{\pi_\theta(o_i|Q)}{\pi_{\theta_{\text{old}}}(o_i|Q)}$ is the likelihood ratio, $A_i$ the advantage, and $\mathrm{D_{KL}}$ is a KL-divergence penalty to keep the policy close to a reference (supervised fine-tuned) model~\cite{ouyang2022training}. In practice, $\epsilon$ and $\beta$ are tuned to balance exploration and stability.  
   
\textbf{Reward Model Design.}
A key differentiator of \OURS is its adherence to a minimal, execution-driven reward formulation. While earlier works have often sought highly elaborate reward signals (aggregating string overlap, schema conformance, partial matching, etc.~\cite{pourreza2025reasoning,ma2025sql}), we find these can encourage reward hacking and brittle behaviors~\cite{skalse2022defining}.  
   
Instead, we define a reward function focused solely on (1) \emph{final execution correctness} and (2) \emph{basic syntax validity}:  
\begin{equation*}  
R =   
\begin{cases}  
1,   & \text{if the execution results exactly align with ground truth;} \\  
0.1, & \text{if syntax is correct and SQL is executable;} \\  
0,   & \text{otherwise.}  
\end{cases}  
\end{equation*}  
Here, execution alignment is determined by running the model's SQL prediction alongside the ground-truth query and matching their results using the strictest available criteria (e.g. BIRD benchmark guidelines), ensuring meaningful progress signals. Syntax validity ensures that models are not unduly penalized for benign formatting or minor structural errors when learning to compose well-formed queries.
Our design is similar to math/logic RL works~\cite{Xie2025LogicRL,guo2025deepseek} with extra constraints on valid SQL execution.

The proposed reward design enables stable, interpretable credit assignment, removing distracting or confounding partial rewards. As empirical results (Section~\ref{sec:results}) demonstrate, our streamlined approach is sufficient and preferable for high-accuracy, generalizable \tts modeling.
\section{Learnings}  
\label{sec:learning}

Our iterative exploration involved experiments with diverse training and data selection strategies, yielding a variety of empirical insights. While we recognize that some of these observations may be context-specific and not universally applicable, we document them here to inform and accelerate future research in the field.

\textbf{Experimental Setup.}  
\looseness=-1 Our primary training datasets are derived from the \bird~\cite{li2024can} training set, and the training/development splits of \spider~\cite{yu2018spider}. Rather than directly adopting these sources, we apply a filtering process aimed at data quality (see Table~\ref{tab:dataset} and Section~\ref{sec:standard_filtering} for details).
\looseness=-1 Initial model selection is guided by performance on the \bird-dev and \spider-test sets—reported via execution accuracy.  As we noticed that a sole focus on these two benchmarks risks overfitting, we later broadened our evaluation suite for a more robust assessment.

As base models, we rely on the Qwen2.5-Coder series~\cite{hui2024qwen2}, including both base, instruct and reasoning-oriented variants.  Unless noted otherwise, GRPO~\cite{shao2024deepseekmath} is used as our default RL algorithm. RL-specific settings include a generation temperature of 0.8, a total batch size of 256 (16 rollouts each), an update batch size of 128 per GRPO step, KL penalty $\beta = 0.001$, and clip ratio  $\epsilon = 0.2$ (see Section~\ref{sec:methodology} for hyperparameter details).
\subsection{Learnings from Training Data}  
\label{sec:learning_from_trianing_data}  

\begin{table}[t]
\caption{
Learning from training data (all with GRPO optimization and Qwen2.5-Coder).
}\centering
\label{tab:data_learning}
\begin{adjustbox}{width=0.99\linewidth}
\centering
\begin{tabular}{clcccccccccccc }
\toprule
{Base Model} & {Training Data} & {\bird-dev} & {\spider-test} \\
\midrule
14B-Inst &\bird, \spider & 64.9 &86.8\\
14B-Inst   &\bird, \spider, \gretel-NonFiltered &64.6 &86.4\\
14B-Inst &\bird, \spider, \gretel-Filtered &\textbf{66.5} &\textbf{88.3}\\
\midrule
32B-Inst   &\bird, \spider, \gretel-Filtered &64.9 &87.7\\
32B-Inst &\bird, \spider,  \gretel-Filtered, \bird-aug-NonFiltered &62.5 &86.1\\
32B-Inst   &\bird, \spider, \gretel-Filtered, \bird-aug-Filtered & 64.9 & 86.8\\
\bottomrule
\end{tabular}
\end{adjustbox}
\end{table}

\textbf{The Critical Role of Filtering.} \label{sec:standard_filtering}  
Thorough inspection of  \bird and \spider training splits reveals that many reference SQL queries in these datasets return empty results upon execution. For RL, where reward signaling is tied to execution correctness, such examples can disrupt the learning process by producing spurious or uninformative rewards. We exclude these and filter out samples with execution times exceeding five seconds, markedly reducing overall training time.  
This straightforward filtering step removed about 1,400 samples from \bird and 1,700 from \spider, yielding a more reliable reward signal and expediting RL convergence. Table~\ref{tab:dataset} provides a summary of datasets used, including derived data splits.  
   
\textbf{Table Enhanced Generation with Model-based Data Filtering for Synthetic Data.}     
We further augment training data with \gretel~\cite{gretel}, which provides schemas without populated data. We use GPT-4o to produce \texttt{INSERT} statements per table (details in \appref{sec:gretel_details}), repeatedly sampling until the reference SQL retrieves non-empty results.
Distractor tables from related domains are randomly added to increase schema complexity, and only queries with SQL length $>160$ characters and successful execution are retained for the non-filtered pool.  
However, naively adding \gretel-NonFiltered to training reduced performance (the first section of~\tref{tab:data_learning}). To remedy this, we employed a model-based filtering---using our best \qwenttb-trained model, we retained only queries where at least one of the ten generations (tempearture=1.0) was correct. This curated \gretel-Filtered set markedly improves results.

\textbf{Unsuccessful Attempts at LLM-Based Data Augmentation.}  
Drawing inspiration from works such as~\cite{yang2024synthesizing,hu-etal-2023-importance}, we attempted to boost diversity through data augmentation, prompting LLMs to paraphrase or generate complex questions given a schema, the original question, and its SQL. Nevertheless, we found that model outputs often closely mirrored the original conditions, limiting diversity. 
We next tried prompting solely with schema descriptions and gold SQL, omitting the original question, and incorporated self-correction~\cite{deng2025reforcetexttosqlagentselfrefinement} to ensure all SQLs were executable and returned data. Despite this, augmented data largely failed to improve model generalization (the second section of~\tref{tab:data_learning}), which we attribute to (1) insufficient linguistic and structural diversity, and (2) oversampling schema-specific patterns, leading to overfitting.  
   
\textbf{Other Data Sources.}\looseness=-1  
We also experimented with model-filtered data from SynSQL-2.5M~\cite{omnisql}, but initial attempts were inconclusive. Given the sheer volume of SynSQL-2.5M, we believe more sophisticated filtering could unlock further gains, which we leave for future work.  
   
\subsection{Learnings from Training Strategy}  
\label{sec:learning_from_trianing_stratety}  
\begin{table}[t]
\caption{
Learnings from training strategy (*: limited results because of the checkpoint loss)
}\centering
\label{tab:strategy_learning}
\begin{adjustbox}{width=0.99\linewidth}
\centering
\begin{tabular}{lcccccccccccccc }
\toprule
{Base Model} & {Training Strategy} & {Optimization} & {\bird-dev} & {\spider-test} \\
\midrule
\qwenttb & --- & \grpo  &64.9 &87.7 \\
\qwenttb & --- & \ppo  &63.0 &85.7\\
\midrule
Qwen2.5-Coder-32B & --- & \grpo &64.4 &87.3 \\
\qwenttb & --- & \grpo &64.9 &87.7 \\
\qwq & --- & \grpo & 55.2 &79.3 \\
\midrule
\qwenttb & Batch RL & \grpo &64.9 &87.7 \\
\qwenttb* & Online RL & \grpo &66.6 & -- \\
\midrule
\qwenttb* & Online RL & \grpo &66.6 & --  \\
\omnisql-32B & Online RL & \grpo &67.9 &88.2  \\
\midrule
\omnisql-32B & Online RL + Self-defined Prompt Template & \grpo &67.9 &88.2\\
\omnisql-32B & Online RL + Modified OmniSQL Prompt& \grpo &70.5	&88.7 \\
\bottomrule
\end{tabular}
\end{adjustbox}
\end{table}  
Unless stated otherwise, the following rely on filtered \bird, \spider, and \gretel.

\textbf{GRPO vs PPO.}  
We compare GRPO and Proximal Policy Optimization (PPO). GRPO, by design, reduces memory usage of critic models and is well-suited for large-scale settings; PPO is simpler and historically popular for stable RL optimization. In our experiments, GRPO outperforms PPO by a comfortable margin (see the first section of~\tref{tab:strategy_learning}), though we caution that PPO may benefit from further hyperparameter tuning.  
   
\textbf{Selecting the Optimal Starting Model.}
\looseness=-1 We assess Qwen-2.5-Coder in the base, instructed, and the reasoning-focused \qwq variant. 
Results show that starting from better instruction following, higher-accuracy models is crucial (see the second section of~\tref{tab:strategy_learning}). 
The main differences between Qwen2.5-Coder-32B base and instruction are (1) the instruction following capability and (2) high-quality instruction finetuning data, which leads to about 0.4 EX gap.
\qwq is optimized for math and reasoning tasks, e.g., math, but not SQL, and its initial accuracy trailed the Qwen-2.5-Coder-instructed version by over 10 points, a gap RL fine-tuning could not close. Later, we discuss how strong supervised checkpoints improve downstream RL. 

\textbf{Online RL Surpasses Batch RL.} 
\looseness=-1 We compare online RL—where the model continually interacts with the environment—with batch RL. Our findings show superior results with online RL, likely due to its increased adaptivity and exposure to more complex negative examples via live interaction (the third section of~\tref{tab:strategy_learning}). This observation extends previous results primarily observed in mathematical and programming, indicating its pertinence for \tts.

\textbf{Supervised Fine-Tuning Model Quality Matters.}
\looseness=-1 Stronger SFT models (e.g., OmniSQL~\cite{omnisql}) consistently yield better downstream RL results (the forth section of~\tref{tab:strategy_learning}), reinforcing the importance of strong initializations and echoing observations from other domains.  
   
\textbf{Prompt Format is Crucial.}  
We observed significant gains (the fifth section of~\tref{tab:strategy_learning}) when switching from a generic prompt to the original \omnisql prompt, adapted for RL training (see~\fref{fig:prompt} for prompt). Prompt structure, inclusion of thinking instructions, and database serialization choices all contributed to improved model performance.  
   
\textbf{Other Observations.} 
We explored variations in rollout count (16, 24, 32), human- or LLM-generated prompts, and various reward designs. None produced significant improvement, and, notably, more fine-grained reward designs induced ``lazy'' behaviors, where models pursued local optima for short-term rewards rather than global correctness.
   
\subsection{Learning from Evaluation Benchmark Diversity}  
\label{sec:learning_from_benchmark}
\begin{table}[t]
\caption{
Diverse evaluation helps to identify generalization capabilities (\omnisql-7B). 
}\centering
\label{tab:eval_learning}
\begin{adjustbox}{width=0.99\linewidth}
\centering
\begin{tabular}{lcccccccccccccc }
\toprule
\multirow{2}{*}{Different setting}& BIRD- & Spider- &Spider2.0- &Spider- & EHR & Science   & \multirow{2}{*}{Average}\\ 
& dev & test &SQLite & DK & SQL &Benchmark  \\
\midrule
\bird only &67.6	&87.8	&8.9	&76.3	&34.9	&50.5	&54.3\\
\bird, \spider, \gretel &67.7	&88.2	&11.9	&79.1	&35.5	&51.8 &55.7\\
\bottomrule
\end{tabular}
\end{adjustbox}
\end{table}

\textbf{Prompt Optimization Benefits OSS Model Evaluation.} 
Prompt selection dramatically affects open-source models (e.g., Llama~\cite{grattafiori2024llama} and Qwen~\cite{hui2024qwen2}) performance, with the \omnisql prompt improving Llama-3.1-70B's \bird-dev accuracy from 57.4\%~\cite{zhai2025excot} to 65.1\% in our experiments. 
Prompt tuning is thus indispensable for the fair comparison of large, general-purpose LLMs in the \tts task.
However, finding the best prompt for general-purpose LLMs is beyond the scope of our work.

\textbf{Diverse Evaluation Surfaces Generalization Gaps.}  
Restricting training to \bird, we observed strong benchmark accuracy on \bird-dev, but an average score 1.4 points lower when evaluated across a broader suite (Table~\ref{tab:eval_learning}), including \bird-dev~\cite{li2024can}, \spider~\cite{yu2018spider}, \spidertwo~\cite{yu2018spider}, \spiderdk~\cite{gan2021exploring}, \ehrsql~\cite{lee2022ehrsql}, and \science~\cite{zhang2023sciencebenchmark}.
This highlights the risk of overfitting and the necessity of evaluating on multiple datasets to ensure robust generalization.

\begin{table}[t]
\setlength{\tabcolsep}{4pt}
\caption{
Comparison between \OURS and other OSS/Proprietary models. *:~SQL-R1~\cite{ma2025sql} uses majority voting in evaluation. **: \cite[Table 2]{pourreza2025reasoning} might use schema linking, our number follows \cite[Table 3/4]{pourreza2025reasoning}.
$^\dagger$: \spidertwo was updated, so we re-evaluate it instead of using the \omnisql number. 
For SQL-specific models, generally sensitive to prompt and database serialization, we take the number from their papers.
DPSK stands for DeepSeek.
}\centering
\label{tab:main_result}
\begin{adjustbox}{width=0.99\linewidth}
\centering
\begin{tabular}{lccccccccccccccc }
\toprule
\multirow{2}{*}{Model} & \multirow{2}{*}{OSS} & \bird & \spider &\spidertwo & Spider & EHR & Science & \multirow{2}{*}{Average} \\
& & (dev) & (test) & -SQLite$^\dagger$ & -DK & SQL & Benchmark \\
\toprule
\multicolumn{9}{c}{\centering \bf{Models Size $<$ 10B}} \\
\midrule
DPSK-Coder-6.7B-Instruct~\cite{guo2024deepseek} & \cmark & 43.1 & 70.5 & 4.4 & 60.9 & 28.6 & 40.8 & 41.4 \\
Qwen2.5-Coder-7B-Instruct~\cite{hui2024qwen2} & \cmark & 50.9 & 82.2 & 4.4 & 67.5 & 24.3 & 45.2 & 45.8 \\
Qwen2.5-7B-Instruct~\cite{yang2024qwen2} & \cmark & 46.9 & 76.8 & 5.2 & 63.7 & 20.9 & 38.5 & 42.0\\
SQL-R1-7B*~\cite{ma2025sql} &\xmark &66.6 &-- &-- &-- &-- &-- &-- \\
\omnisql-7B~\cite{omnisql} & \cmark & 63.9 & 87.9 &13.3  & 76.1 & 34.9 & 50.2 & 54.4\\
Think2SQL-7B~\cite{papicchio2025think2sqlreinforcellmreasoning} &\xmark &56.1 &-- &-- &-- &-- &-- &-- \\
OpenCoder-8B-Instruct~\cite{huang2024opencoder} & \cmark & 37.5 & 68.3 & 1.5 & 62.6 & 21.9 & 39.8 & 38.6\\
Meta-Llama-3.1-8B-Instruct~\cite{grattafiori2024llama} & \cmark & 42.0 & 72.2 &1.5 & 62.6 & 24.6 & 36.8 & 40.0 \\
Granite-8B-Code-Instruct~\cite{mishra2024granite} & \cmark & 27.6 & 64.9 &1.5 & 50.7 & 16.0 & 29.4 &31.7  \\
Granite-3.1-8B-Instruct~\cite{mishra2024granite} & \cmark & 36.0 & 69.8 &3.7 & 60.0 & 19.6 & 36.8 &37.7 \\
\midrule
\oursb & \cmark &\textbf{68.9} &\textbf{88.8} &\textbf{15.6} &\textbf{81.5} &\textbf{36.7} &\textbf{51.8}	&\textbf{57.2}\\

\toprule
\multicolumn{9}{c}{\centering \bf{10B $\leq$ Models Size $\leq$ 30B}} \\
\midrule
Qwen2.5-Coder-14B-Instruct~\cite{hui2024qwen2} & \cmark & 61.5 & 86.6 & 11.1 & 73.6 & 31.6 & 52.2 & 52.8\\
Qwen2.5-14B-Instruct~\cite{yang2024qwen2} & \cmark & 56.7 & 82.0 &8.1 & 72.3 & 28.8 & 51.2 &49.9 \\
\omnisql-14B~\cite{omnisql} & \cmark & 64.2 & 88.3 &12.6 & 72.9 & 39.9 & 56.9 & 55.8\\
Reasoning-SQL-14B**~\cite{pourreza2025reasoning} &\xmark &64.2 &81.4 &-- &73.0 &-- &-- &-- \\
Starcoder2-15B-Instruct~\cite{lozhkov2024starcoder} & \cmark & 38.5 & 73.0 &1.5  & 66.5 & 16.8 & 25.8 & 37.0\\
DPSK-Coder-V2-Inst (16B/MoE)~\cite{zhu2024deepseek} & \cmark & 44.6 & 77.9 &2.2 & 63.7 & 23.9 & 39.1 & 41.9\\
Granite-20B-Code-Instruct~\cite{mishra2024granite} & \cmark & 34.0 & 74.1 & 1.5 & 62.2 & 23.5 & 37.5 & 38.8 \\
Codestral-22B~\cite{codestral2024} & \cmark & 52.7 & 78.6 &8.1 & 69.9 & 37.8 & 48.5 & 49.3 \\
\midrule
\ourfb &\cmark &\textbf{70.1}	&\textbf{89.4}	&\textbf{16.3}	&\textbf{79.4} &\textbf{40.7} &\textbf{58.2} &\textbf{59.0}\\

\toprule
\multicolumn{9}{c}{\centering \bf{30B $<$ Models Size or Unknown}} \\
\midrule
Qwen2.5-Coder-32B-Instruct~\cite{hui2024qwen2} & \cmark & 64.5 & 87.5 &10.4 & 78.3 & 36.4 & 54.8 & 55.3 \\
Qwen2.5-32B-Instruct~\cite{yang2024qwen2} & \cmark & 62.0 & 84.9 &10.4  & 73.1 & 33.6 & 50.5 & 52.4 \\
Xiyan-SQL-32B~\cite{gao2024xiyan} &\cmark &67.0 &-- &-- &-- &-- &-- &--\\
ExDPO-32B~\cite{zhai2025excot} &\cmark &68.3 &85.1 &-- &-- &-- &-- &--\\
\omnisql-32B~\cite{omnisql} & \cmark & 64.5 & 87.6 & 14.8 & 76.1 & 42.4 & 57.2 & 57.1  \\
DPSK-Coder-33B-Instruct~\cite{guo2024deepseek} & \cmark & 49.2 & 74.3 &5.2  & 69.0 & 31.4 & 44.5 &45.6 \\
Granite-34B-Code-Instruct~\cite{mishra2024granite} & \cmark & 33.8 & 74.4 & 0.0 & 64.7 & 23.8 & 40.1 & 39.5\\
Mixtral-8x7B-Inst. (47B, MoE)~\cite{jiang2024mixtral} & \cmark & 35.3 & 67.8 &3.7 & 55.3 & 21.5 & 29.4 & 35.5\\
Meta-Llama-3.1-70B-Instruct~\cite{grattafiori2024llama} & \cmark & 65.1 & 84.3 & 7.4 & 75.1 & 37.4 & 55.2 & 54.1\\
ExDPO-70B~\cite{zhai2025excot} &\cmark &68.5 &86.6 &-- &-- &-- &-- &--\\
Qwen2.5-72B-Instruct~\cite{yang2024qwen2} & \cmark & 60.3 & 84.0 & 11.9 & 76.4 & 35.0 & 52.8 & 53.4 \\
Command-A-111B~\cite{cohere2025commanda} & \cmark &63.5&-- &-- &-- &-- &-- &--\\
DeepSeek-V3 (671B, MoE)~\cite{liu2024deepseek} & \cmark & 63.2 & 85.5 &14.8  & 72.9 & 43.2 & 56.2 & 55.6 \\
GPT-4o-mini~\cite{openai2023gpt4} & \xmark & 58.8 & 82.4 & 11.9 & 73.3 & 37.9 & 51.8 & 52.7\\
GPT-4-Turbo~\cite{openai2023gpt4} & \xmark & 62.0 & 83.4 & 13.3 & 72.3 & 43.1 & 59.2 & 55.6\\
GPT-4o~\cite{openai2023gpt4} & \xmark & 61.9 & 83.2 & \textbf{17.0} & 72.9 & \textbf{44.9} & 55.5 & 55.9\\
\midrule
\ourttb &\cmark &\bf{70.5}	&\textbf{88.7}	&	16.3 &\textbf{80.6}	&40.1	&\textbf{60.9}	&\textbf{59.5}\\

\bottomrule
\end{tabular}
\end{adjustbox}
\end{table}

\section{Main Result}
\label{sec:results}  

Our final experiments build upon all prior empirical insights: the training data includes the filtered \bird-training and \spider-training/dev sets, along with the model-based filtered \gretel-Filtered synthetic examples. 
The reinforcement learning setup consists of (1) GRPO as the optimization algorithm, (2) online training, (3) initializing from \omnisql supervised checkpoints, and (4) using a modified \omnisql prompt for both training and evaluation. 
We report execution-based accuracy results across all six benchmarks detailed in~\tref{tab:dataset}.

\subsection{Single-Model, Single-Inference BIRD Test Benchmark Performance}  
\label{sec:single_model_perf}  

\OURS achieves new SOTA record on \bird Test benchmarks and the benchmark numbers are reported in~\tref{tab:bird_leaderboard}, which summarizes the performance of top models on the BIRD benchmark under strict single-model conditions.
Note that \OURS is further constrained by single-inference (a.k.a. greedy decoding) evaluation to reflect real-world deployment needs.\footnotemark[1]
\OURS 14B and 32B both surpass the 70\% accuracy threshold on the BIRD test split, setting a new standard.
Specifically, \OURS-32B achieves 71.83\%—exceeding the previous best, XiYan-32B, by 2.8 points. 
Furthermore, \OURS-7B matches the performance of previous strong models such as ExCoT-70B~\cite{zhai2025excot} with $10\times$ fewer parameters.

\subsection{Performance Evaluation across Six Benchmarks}  

Instead of optimizing the model for single benchmark performance, \OURS achieves best-in-class average performance across six different benchmarks.
Table~\ref{tab:main_result} presents a comprehensive comparison between \OURS and a wide range of open-source and proprietary models, including general-purpose LLMs (e.g. DeepSeek-V3, GPT-4o, GPT-4-Turbo), coding models (Qwen-Coder), and specialized \tts models (XiYan-SQL, OmniSQL, ExCoT, SQL-R1, Reasoning-SQL). 
Except where otherwise indicated, all evaluations use single-model, single-inference (i.e., greedy decoding with no ensembling, schema linking, or external re-ranking).

\textbf{State-of-the-Art Across All Scales.}    
\OURS consistently outperforms existing state-of-the-art models across all parameter scales and benchmarks. For models under 10B parameters, \oursb achieves an average accuracy of 57.2, outperforming open-source competitors (e.g., \omnisql-7B at 54.4) and closely rivaling, or surpassing, closed-source models such as SQL-R1-7B.    
In the 10B–30B range, \ourfb attains 59.0, exceeding strong baselines like \omnisql-14B and Reasoning-SQL-14B. In the 30B+ category, \ourttb achieves the best results overall, with an average of 59.5, outperforming the largest open-source and commercial models, including DeepSeek-V3 (55.6), GPT-4-Turbo (55.6), and even GPT-4o (55.9).  

\begin{table}[t]
\begin{minipage}[t]{.545\linewidth}
    \centering
    \includegraphics[width=0.99\textwidth]{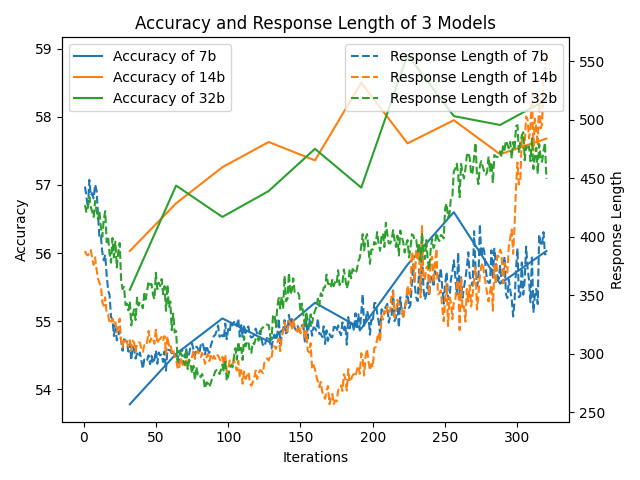}
    \captionof{figure}{Generation length and the average accuracy across six benchmarks.}
    \label{figure:length}
\end{minipage}\hspace{1em}\hfill
\begin{minipage}[t]{.455\linewidth}
\vspace{-4.5cm}
    \centering
    \captionof{table}{\OURS-14B (top) and 32B (bottom) with different extra evaluation techniques on \bird-dev.}
    \label{tab:combine_inference_technique}
    \begin{adjustbox}{width=0.59\linewidth}
    \begin{tabular}{ccc}
        \toprule
        Value & Majority & \bird \\
        Retrieval & Voting & -dev \\
        \midrule
        \xmark & \xmark & 70.1 \\
        \cmark & \xmark & 70.5 \\
        \xmark & \cmark & 70.6 \\
        \cmark & \cmark & 70.8 \\
        \midrule
        \xmark & \xmark & 70.5 \\
        \cmark & \xmark & 70.9 \\
        \xmark & \cmark & 71.2 \\
        \cmark & \cmark & 71.5 \\
        \bottomrule
    \end{tabular}
    \end{adjustbox}
\end{minipage}
\vspace{-6mm}
\end{table}

\textbf{Parameter Efficiency and Task Specialization.}    
Notably, \OURS-7B matches or surpasses models such as DeepSeek-V3 (671B, MoE) and \omnisql-32B  on six benchmark accuracy, despite a fraction of their parameter count. These results highlight the advantages of task-specific training and reward design for compositional SQL reasoning.  
   
\textbf{Benchmark-Specific Highlights.}    
On challenging benchmarks such as BIRD, Spider, and EHRSQL, \OURS shows marked improvements. For instance, on the BIRD-dev split, \OURS 7B, 14B, and 32B reach 68.9, 70.1, and 70.5, setting new records across size categories. 
Substantial gains are also seen on Spider-DK and the Science benchmark, underscoring \OURS's robustness and domain coverage.  
\section{Discussion}
Beyond the benchmark achievements, the development of \OURS revealed some insights into model behavior and potential for exploiting inference-time enhancements. These also motivate further examination of the model's comparative standing.

\textbf{Enhancing \OURS with Other Techniques.}
While \OURS is designed for efficient, direct inference, it can also serve as the SQL generator in more complex agentic systems. Table~\ref{tab:combine_inference_technique} shows results using two inference-time enhancements: value retrieval (from \cite{talaei2024chess}) and majority voting (as in \cite{omnisql,ma2025sql}) with eight generations per sample. Each technique yields a tangible improvement; combined, they boost \OURS-32B performance by up to one point on BIRD-dev, demonstrating complementary gains from orthogonal reasoning components.  
   
\textbf{Response Length vs. Accuracy.}   
Figure~\ref{figure:length} plots the evolution of average response length and execution accuracy across training for all three sizes of \OURS. We observe a U-shaped pattern in response length—initially decreasing, then gradually increasing—as accuracy rises.
This trend may reflect early-stage conservative decoding (and possible undertraining), followed by richer, more grounded outputs as training progresses, an effect also noted in some of the previous works \cite{Xie2025LogicRL}.

To further validate the efficacy of \OURS, we contrast its performance with recently published, competitive approaches in SQL generation.

\begin{table}[t]
\caption{
The comparison between \OURS and Reasoning-SQL~\cite{pourreza2025reasoning}. The numbers for the latter from~\cite[Tables 4 and 5]{pourreza2025reasoning} and it might involve schema linking as mentioned in~\cite[Table 2]{pourreza2025reasoning}, but we are unsure; see~\tref{tab:techniques_comparison} for the reward design comparison.
}\centering
\label{tab:reasoningsql_comparison}
\begin{adjustbox}{width=0.9\linewidth}
\centering
\begin{tabular}{lccccccccccccc }
\toprule
\multirow{2}{*}{Method} & \multirow{2}{*}{Base Model} & \multirow{2}{*}{Filtering Method} & Reward & \bird & \spider &\spider \\
& & & Design & -dev & -test & -DK \\
\midrule
Reasoning-SQL & \qwenfb &Large Models Prompt& Complex & 64.21 &81.43 &73.03 \\
\OURS & \qwenfb & Empty Golden-SQL & Simple &66.49 &87.20 &75.10 \\
\bottomrule
\end{tabular}
\end{adjustbox}
\end{table}
\begin{table}[t]
\caption{
Comparison of \OURS, SQL-R1~\cite{ma2025sql} and Think2SQL~\cite{papicchio2025think2sqlreinforcellmreasoning}. 
GRE: greedy decoding, MV: Majority Voting (8 candidates). 
Competitors numbers according to \cite[Table 2]{ma2025sql} and ~\cite[Table 1 and 3]{papicchio2025think2sqlreinforcellmreasoning}; see~\tref{tab:techniques_comparison} for the reward design comparison.
}\centering
\label{tab:sqlrone_comparison}
\centering
\begin{adjustbox}{width=0.9\linewidth}
\begin{tabular}{lcccccccccccccc }
\toprule
\multirow{2}{*}{Method} & \multirow{2}{*}{Base Model} & SFT & Reward & \bird-dev & \spider-test  \\
& & (\#Samples) & Design & GRE/MV & GRE/MV \\
\midrule
SQL-R1 & \qwensb &\xmark & Complex & --/63.1  &--/86.1 \\
SQL-R1 & \qwensb &\cmark (200K) & Complex & --/59.2  &--/86.4 \\
Think2SQL &\qwensb & \cmark (9K) & Complex &56.1/-- & 82.4/-- \\
\OURS & \qwensb &\xmark & Simple &\textbf{63.9}/\textbf{64.8} &\textbf{85.0}/\textbf{87.1} \\
\midrule
SQL-R1 & \omnisqlsb &\xmark  & Complex & --/66.6 &--/\bf{88.7} \\
\OURS & \omnisqlsb &\xmark  & Simple &\textbf{67.6}/\textbf{69.4} &\textbf{87.8}/88.6 \\
\bottomrule
\end{tabular}
\end{adjustbox}
\end{table}  

\textbf{Reasoning-SQL.}  
\looseness=-1 To enable a fair comparison with Reasoning-SQL~\cite{pourreza2025reasoning}, we retrain \OURS from the Qwen-2.5-14B-Instruct base, use BIRD-training data alone, and run 3 epochs. Under these controlled settings, the two methods differ primarily in training data filtering (\OURS uses simple empty-return-based removal, Reasoning-SQL uses LLM filtering) and reward design (our simple execution/syntax signal vs. a complex mixture).  
As shown in~\tref{tab:reasoningsql_comparison}, \OURS outperforms Reasoning-SQL by clear margins on BIRD-dev (+2.28), Spider-test (+5.77), and Spider-DK (+2.07). This underscores the power of direct, execution-guided reinforcement learning and streamlined filtering.  

\textbf{SQL-R1 and Think2SQL.}
\looseness=-1 We further compare \OURS with SQL-R1~\cite{ma2025sql} and Think2SQL~\cite{papicchio2025think2sqlreinforcellmreasoning}, restricting RL fine-tuning to BIRD-training and evaluating under both greedy and majority voting regimes. As summarized in Table~\ref{tab:sqlrone_comparison}, \OURS delivers the best or near-best performance in all configurations and consistently outperforms prior approaches, even without extensive supervised pretraining or complex reward engineering.  
\section{Conclusions}
\label{sec:conclusions}
\looseness=-1 We presented \OURS, a novel RL framework for \tts that uses GRPO and a simple execution-based reward.
It achieves 71.83\% execution accuracy on BIRD-test as a single 32B model—setting a new state-of-the-art and surpassing much larger LLMs. 
Across six challenging benchmarks, \OURS yields up to 4 points improvement over strong baselines, with notable parameter efficiency (e.g., the 7B version matches or outperforms prior ExCoT-70B models and GPT4o).

Our results highlight that minimal reward signals, principled data filtering, and carefully curated training strategies are key for robust SQL generation. We release models to support future research in this area.

{
\bibliographystyle{plain}
\bibliography{ref.bib}
}

\clearpage
\onecolumn
\appendix

\counterwithin{figure}{section}
\counterwithin{table}{section}

\section{Construction Pipeline from \gretel to \gretel-NonFiltered}
\label{sec:gretel_details}

To enable SQL execution and ensure meaningful query results, we use a carefully designed prompt to guide an LLM in generating realistic table data. Appendix~\ref{fig:gretel_prompt} illustrates this prompt, which produces valid \texttt{INSERT} statements aligned with both the SQL context and query.

\begin{figure}[!ht]
\begin{prompt}{Prompt for Adding Synthetic Data}
\textbf{Adding Synthetic Data Prompt}\\
You are an expert in SQL data modeling. Your task is to analyze the given SQL schema and, if necessary, generate realistic and logically consistent sample data to ensure:\\
For a given \textless SQL Prompt\textgreater, both \textless SQL Query\textgreater\ and \textless SQL Context\textgreater\ can meet its requirements, and \textless SQL Query\textgreater\ can query the corresponding data from the TABLE created by \textless SQL Context\textgreater.\\

Given a -- \verb|**|\textless SQL Prompt\textgreater\verb|**|:\\
\{question\}\\

I have generated the \textless SQL Query\textgreater\ and \textless SQL Context\textgreater:\\
-- \verb|**|\textless SQL Query\textgreater\verb|**|:\\
\texttt{\{sql\_query\}}\\

-- \verb|**|\textless SQL Context\textgreater\verb|**|:\\
\texttt{\{sql\_context\}}\\

\{error\_infomations\}\\

I need data samples to validate the correctness of the \textless SQL Query\textgreater.\\
Therefore, please help me add one \texttt{INSERT} statement for each table in the \textless SQL Context\textgreater, with 5 sample rows per table.\\
The inserted data should ensure that the \textless SQL Query\textgreater\ can retrieve results from the tables.\\
Please ensure that it does not cause errors when using \texttt{sqlite3}.\\
Please do not include any additional explanations or instructions.\\

Please help me fix this \verb|**|\textless SQL Context\textgreater\verb|**| and ensure that it contains at most five records.\\
Please also help me modify \verb|**|\textless SQL Query\textgreater\verb|**| to ensure that it does not cause errors when using \texttt{sqlite3}.\\

Please give your expanded \verb|**|\textless SQL Context\textgreater\verb|**| in:
\texttt{\textbackslash\textbackslash sql\_context}\\
your fixed \verb|**|\textless SQL Query\textgreater\verb|**| in:
\texttt{\textbackslash\textbackslash sql\_query}\\
and the \verb|**|INSERT statements\verb|**| in:
\texttt{\textbackslash\textbackslash sql\_insert}
\end{prompt}
\caption{Prompt for Generating Executable SQL Context and Synthetic Data Inserts in the \gretel Pipeline}
\label{fig:gretel_prompt}
\end{figure}

These synthetic data inserts serve as the foundation for the construction pipeline outlined in Algorithm~\ref{alg:gretel_nonfiltered_construction}, which augments each sample with distractive tables and applies filtering to retain executable, informative SQL examples.

\begin{algorithm}[!ht]
\caption{Construction of \texttt{\gretel-NonFiltered}: Inserting Synthetic Data, Adding Distractive Tables, and Filtering Executable SQL Samples}
\label{alg:gretel_nonfiltered_construction}
\KwIn{Full dataset \texttt{\gretel}}
\KwOut{\texttt{\gretel-NonFiltered} after adding synthetic data and distractive tables}

\textbf{1. Insert Data Using LLM}\\

Initialize \texttt{\gretel-NonFiltered} as empty list\;

\ForEach{\texttt{sample} in \texttt{\gretel}}{
    Initialize \texttt{\gretel-ForNextRound} $\gets$ \texttt{[sample]}\;

    \For{$i \gets 1$ \KwTo $8$}{
        Initialize \texttt{next\_round} as empty list\;

        \ForEach{\texttt{sample} in \texttt{\gretel-ForNextRound}}{
            \tcc{We use GPT-4o to generate insert statements; prompt is provided in Appendix~\ref{fig:gretel_prompt}}
            \texttt{insert\_stmts} $\gets$ \texttt{GPT\_generate\_inserts(sample)}\;

            Execute \texttt{sql\_context} to create tables\;
            Execute \texttt{insert\_stmts} to populate data\;

            \texttt{results} $\gets$ \texttt{execute\_sql\_query(sample.sql\_query)}\;

            \eIf{\texttt{results} not empty}{
                Append \texttt{sample} to \texttt{\gretel-NonFiltered}\;
            }{
                Append \texttt{sample} to \texttt{next\_round}\;
            }
        }

        \texttt{\gretel-ForNextRound} $\gets$ \texttt{next\_round}\;
    }
}

\vspace{0.5em}
\textbf{2. Add Distractive Tables}\\
\ForEach{\texttt{sample} in \texttt{\gretel-NonFiltered}}{
    \texttt{schema} $\gets$ \texttt{sample.sql\_context}\;
    \texttt{domain} $\gets$ \texttt{sample.domain}\;
    \texttt{existing\_tables} $\gets$ \texttt{extract\_table\_names(schema)}\;
    \If{\texttt{existing\_tables} is None}{
        \textbf{continue}\;
    }

    \tcc{Sample table count from BIRD/Spider distribution with added uniform noise}
    \texttt{n} $\gets$ \texttt{sample\_with\_noise\_from\_table\_distribution(table\_counts)}\;

    \tcc{Select non-conflicting table schemas from the same domain}
    \texttt{distractive\_schemas} $\gets$ \texttt{select\_nonconflicting\_schemas(domain, existing\_tables, n - 1)}\;

    \tcc{Create distractive tables first, then target tables which may overwrite on name conflict}
    \texttt{create\_database(distractive\_schemas + schema)}\;
}

\vspace{0.5em}
\textbf{3. Final Selection}\\
\ForEach{\texttt{sample} in \texttt{\gretel-NonFiltered}}{
    \If{length of \texttt{sample.sql\_query} $>$ 160 \textbf{and} \texttt{sample.sql\_query} returns results}{
        Keep \texttt{sample} in \texttt{\gretel-NonFiltered}\;
    }
    \Else{
        Discard \texttt{sample} from \texttt{\gretel-NonFiltered}\;
    }
}

\Return{\texttt{\gretel-NonFiltered}}\;
\end{algorithm}

\section{Data Augmentation}
\label{app:Augmentation}

\begin{figure}
\begin{prompt}{Prompt for BIRD Data Augmentation}
Table information: \\
Table name: \{table name\} \\
Column name: \{column name\} \\
Column description: \{column desc\} \\
Sample rows: \{samples\} \\
\\
Optional(Task: \{task\}. The answer to the task is: \{answer\}.)\\
\\
Based on this, write 10 more complex nested SQLite SQL queries or SQLs with CTEs in sql code block format. You can use any information in the database information provided. Each query should be different. You can write SELECT query only. For each query, just write one sentence to describe the task. Format like:\\
\\
/*Task: \{task description in one sentence\}*/\\
SELECT ...\\
\\
Don't output other contents.
\end{prompt}
\caption{Prompt for BIRD Data Augmentation}
\label{fig:data_aug_prompt}
\end{figure}

\begin{algorithm}[!ht]
\caption{Self-Correction Workflow: Execution of multiple SQLs with self-correction and refinement based on invalid result feedback to ensure augmented SQLs with valid results.}
\label{alg:self_correction_workflow}
\KwIn{List of SQL statements \texttt{sqls}}
\KwOut{List of successful results \texttt{result\_dic\_list}}

Initialize \texttt{result\_dic\_list} as empty list\;
Initialize \texttt{error\_rec} as empty list\;

\While{\texttt{sqls} not empty}{
    \texttt{sql} = \texttt{sqls[0]}\;
    \texttt{results} = \texttt{execute\_sql\_sqlite(sql)}\;

    \If{\texttt{results} is valid (i.e., string and not empty)}{
        Append \texttt{sql} and \texttt{results} to \texttt{result\_dic\_list}\;
        Continue to next SQL\;
    }

    Initialize \texttt{max\_try}\;
    \While{\texttt{results} is not valid}{
        \If{\texttt{max\_try == 0}}{
            \textbf{break}\;
        }
        \texttt{corrected\_sql} $\gets$ \texttt{self\_correct(sql, results)}\;
        \If{\texttt{corrected\_sql} is not valid}{
            \textbf{continue}\;
        }
        Execute corrected SQL; \texttt{results} = \texttt{execute\_sql\_sqlite(sql)}\;
        Decrease \texttt{max\_try} by 1;
    }

    \If{\texttt{results} is valid}{
        \If{\texttt{sqls} not empty}{
            \texttt{sqls} $\gets$ \texttt{similar\_error\_refine(sqls)}\;
        }

        \If{\texttt{corrected\_sql} exists}{
            Append \texttt{corrected\_sql} and \texttt{results} to \texttt{result\_dic\_list}\;
        }
    }
}
\Return{\texttt{result\_dic\_list}}\;

\end{algorithm}

\begin{figure}[!ht]
\begin{prompt}{Prompt for Self-Correction Workflow}
\textbf{Self-Correction Prompt}\\
Input SQL: \{sql\}\\
The error information is: \{error\}\\
Please correct the SQL based on the previous context. Output your reasoning process followed by only one corrected SQL query in the following format:\\
-- Description: ...\\
<Corrected SQL here>\\
Do not output multiple SQLs or only an analysis without a final SQL.\\
\\
\textbf{Similar Error Refinement Prompt}\\
The following SQL has been corrected:\\
Original SQL: \{sql\}\\
Corrected SQL: \{corrected\_sql\}\\
Please correct the remaining SQL statements if they contain similar errors. The list of SQLs to be refined is: \{sqls\}\\
For each corrected SQL, respond in the following format:\\
-- Description: ...\\
<Corrected SQL here>
\end{prompt}
\caption{Prompt for Self-Correction Workflow}
\label{fig:self_correct_prompts}
\end{figure}

We perform data augmentation on the BIRD dataset following the prompt template shown in Figure~\ref{fig:data_aug_prompt}. Note that the \textit{task} component is optional. To encourage diversity in the generated data, we omit the \textit{task} and \textit{answer} fields when necessary. 

After obtaining a set of SQL queries, we apply a self-correction workflow as described in Algorithm~\ref{alg:self_correction_workflow} to ensure the SQLs are executable and yield non-empty results. The prompts used for \texttt{self\_correct} and \texttt{similar\_error\_refine} are provided in Figure~\ref{fig:self_correct_prompts}.

\section{Prompt Used For Training/Evaluation}
We present the prompt used for our training and evaluation in~\fref{fig:prompt}.

\begin{figure}[!ht]
    \centering
    \begin{prompt}{Prompt Template for Training/Evaluation}
\textbf{System:}\newline
You are a data science expert. Below, you are provided with a database schema and a natural language question. Your task is to understand the schema and generate a valid SQL query to answer the question.\\
\newline
\textbf{User:}\\
Database Engine:\\
SQLite\\
\newline
Database Schema:
\{Database Schema\}\\
This schema describes the database's structure, including tables, columns, primary keys, foreign keys, and any relevant relationships or constraints.\\
\\
Question:\\
\{evidence + question\}\\
\\
Instructions:\\
- Make sure you only output the information that is asked in the question. If the question asks for a specific column, make sure to only include that column in the SELECT clause, nothing more.\\
- The generated query should return all of the information asked in the question without any missing or extra information.\\
- Before generating the final SQL query, please think through the steps of how to write the query.\\

Output Format:\\
Please provide a detailed chain-of-thought reasoning process and include your thought process within `<think>` tags. Your final answer should be enclosed within `<answer>` tags.\\
\newline
Ensure that your SQL query follows the correct syntax and is formatted as follows:\\
\newline
\verb|```|sql\newline
-- Your SQL query here\newline
\verb|```|\newline
\newline
Example format:\newline
<think> Step-by-step reasoning, including self-reflection and corrections if necessary. [Limited by 4K tokens] </think>\newline
<answer> Summary of the thought process leading to the final SQL query. [Limited by 1K tokens]\newline
\newline
\verb|```|sql\newline
Correct SQL query here\newline
\verb|```|\newline
</answer>\\
\newline
\textbf{Assistant:}\\
Let me solve this step by step.\\
<think>
\end{prompt}
\caption{Prompt Template for Training and Evaluation}
\label{fig:prompt}
\end{figure}

\section{Examples from \OURS-32B}
We provide several examples here to illustrate the thinking process of \OURS-32B in~\fref{fig:example_1}, \ref{fig:example_2}, and~\ref{fig:example_3}. 
Note that we did not cherry-pick the answer so the generated SQL might be wrong.

\begin{figure}[!ht]
    \centering
    \begin{prompt}{Example I of response from \OURS-32B}

Question: Eligible free rate for K-12 = `Free Meal Count (K-12)` / `Enrollment (K-12)`\\What is the highest eligible free rate for K-12 students in the schools in Alameda County?\\
\\
\textbf{Assistant:}\\
Let me solve this step by step.\\
<think> \\
1. **Identify the goal**: We need to find the highest eligible free rate for K-12 students in the schools located in Alameda County. The rate is calculated as `Free Meal Count (K-12)` divided by `Enrollment (K-12)`.\\
\\
2. **Determine the relevant table and columns**: \\   - The relevant table is `frpm` because it contains the columns `Free Meal Count (K-12)` and `Enrollment (K-12)`.\\   - However, we need to filter by County, which is in the `schools` table. Thus, we will need to join the `frpm` table with the `schools` table on the `CDSCode` column to filter by the county name.\\\\3. **Formulate the query**:\\   - We will calculate the eligible free rate by taking the ratio of `Free Meal Count (K-12)` and `Enrollment (K-12)`.\\   - The `schools` table will be used to filter for schools in Alameda County.\\   - The `MAX` function will be used to find the highest eligible free rate.\\\\4. **Join tables**:\\   - Use an `INNER JOIN` between the `schools` table and the `frpm` table on the `CDSCode` column to link them correctly.\\</think>\\\\<answer>\\Here is the correct SQL query:\\\\\verb|```|sql\\SELECT MAX(f.`Free Meal Count (K-12)` / f.`Enrollment (K-12)`) AS HighestEligibleFreeRate\\FROM frpm AS f\\JOIN schools AS s ON f.CDSCode = s.CDSCode\\WHERE s.County = 'Alameda';\\\verb|```|\\\\**Explanation**:\\- The `JOIN` is used to connect the `frpm` and `schools` tables on the `CDSCode` column.\\- The `WHERE` clause filters the records to only include those from Alameda County.\\- The `MAX` function is used to find the highest value of the eligible free rate, which is calculated as `Free Meal Count (K-12)` divided by `Enrollment (K-12)`.\\\\The resulting query is efficient and meets the requirements of the natural language question.\\</answer>
\end{prompt}
\caption{Example I of response from \OURS-32B}
\label{fig:example_1}
\end{figure}

\begin{figure}[!ht]
    \centering
    \begin{prompt}{Example II of response from \OURS-32B}

Question: Valid charter number means the number is not null. Rank schools by their average score in Writing where the score is greater than 499, showing their charter numbers.\\
\\
\textbf{Assistant:}\\
Let me solve this step by step.\\
<think>\\
1. **Understanding the Question**:\\   - We need to rank schools based on their average score in writing (`AvgScrWrite`) for scores greater than 499.\\   - We should display the `CharterNum` of the school.\\   - A valid charter number means the number is not null.\\\\2. **Identifying the Necessary Tables**:\\   - The `satscores` table contains the average scores for writing (`AvgScrWrite`) and the `cds` code which can be used to join with the `schools` table.\\   - The `schools` table contains the `CharterNum` field which holds the charter number of the schools.\\   - We need a join between the `satscores` and `schools` tables on the `cds` (from `satscores`) and `CDSCode` (from `schools`) fields.\\\\3. **Filter Conditions**:\\   - We need to filter out schools with `AvgScrWrite` greater than 499.\\   - The `CharterNum` should not be null.\\\\4. **Sorting**:\\   - The result needs to be sorted by the average score in writing (`AvgScrWrite`), in descending order.\\\\5. **Selecting Columns**:\\   - We are only interested in the `CharterNum` field from the `schools` table.\\\\</think>\\\\<answer>\\\\\verb|```|sql\\SELECT T2.CharterNum\\FROM satscores AS T1\\INNER JOIN schools AS T2\\ON T1.cds = T2.CDSCode\\WHERE T1.AvgScrWrite > 499 AND T2.CharterNum IS NOT NULL\\ORDER BY T1.AvgScrWrite DESC;\\\verb|```|\\\\</answer>\\\\This query will:\\- Join the `satscores` and `schools` tables based on the `cds` and `CDSCode` fields.\\- Filter out schools with an average writing score (`AvgScrWrite`) greater than 499 and where `CharterNum` is not null.\\- Sort the results by `AvgScrWrite` in descending order.\\- Return only the `CharterNum` field.
\end{prompt}
\caption{Example II of response from \OURS-32B}
\label{fig:example_2}
\end{figure}

\begin{figure}[!ht]
    \centering
    \begin{prompt}{Example III of response from \OURS-32B}

Question: Eligible free rates for students aged 5-17 = `Free Meal Count (Ages 5-17)` / `Enrollment (Ages 5-17)`\\What is the free rate for students between the ages of 5 and 17 at the school run by Kacey Gibson? \\
\\
\textbf{Assistant:}\\
Let me solve this step by step.\\
<think>\\
1. The question asks for the rate of students between the ages of 5 and 17 who are eligible for free meals. The formula provided for this rate is `Free Meal Count (Ages 5-17) / Enrollment (Ages 5-17)`.\\2. From the database schema provided, the relevant data can be found in the `frpm` table, which contains the columns `Free Meal Count (Ages 5-17)` and `Enrollment (Ages 5-17)`.\\3. The condition for the school is based on an administrator's name, Kacey Gibson. This information is stored in the `schools` table in the columns `AdmFName1` and `AdmLName1`.\\4. To link the `frpm` table to the `schools` table, we use the `CDSCode` column, which is the primary key in both tables.\\5. The final query needs to join the `frpm` and `schools` tables on their common `CDSCode` key, filter the rows for the specific administrator, and then compute and return the eligible free rate for students aged 5-17.\\</think>\\<answer>\\\verb|```|sql\\SELECT \\    (T1.`Free Meal Count (Ages 5-17)` / T1.`Enrollment (Ages 5-17)`) AS free\_rate\\FROM \\    frpm AS T1\\JOIN \\    schools AS T2\\ON \\    T1.CDSCode = T2.CDSCode\\WHERE \\    T2.AdmFName1 = 'Kacey' AND T2.AdmLName1 = 'Gibson';\\\verb|```|\\This SQL query is correct and directly answers the question with the necessary steps and logic based on the provided schema.\\</answer>
\end{prompt}
\caption{Example III of response from \OURS-32B}
\label{fig:example_3}
\end{figure}

\section{BIRD Single Model Board Screenshot}
\begin{figure}
    \centering
    \includegraphics[width=0.99\linewidth]{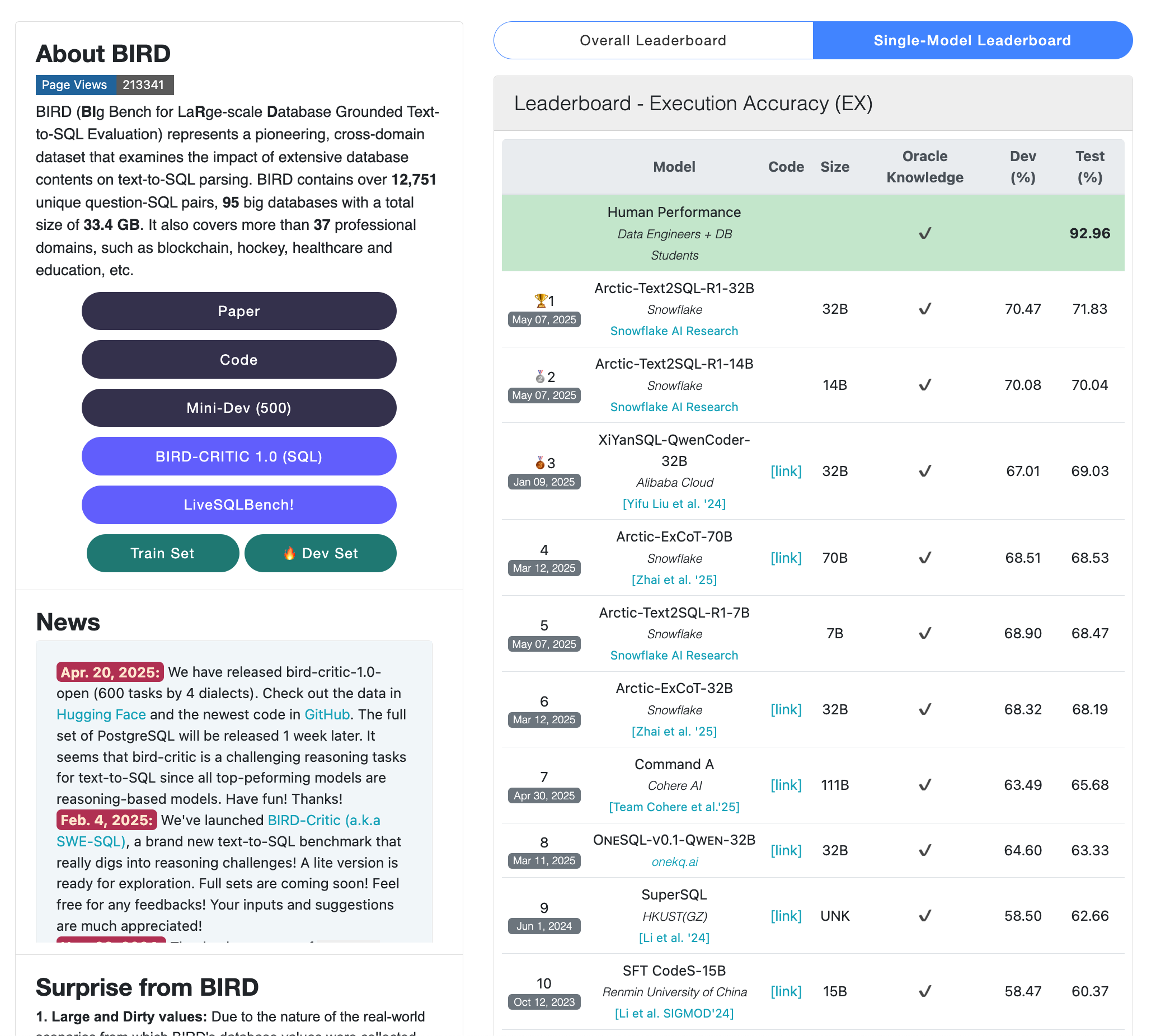}
    \caption{BIRD Single Model Board Screenshot on May 22nd 2025}
    \label{fig:bird_leaderboard}
\end{figure}

\section{Compute Resource}
\label{sec:compute_resource}
All experiments were conducted on 8 H100-80GB nodes (a total of 64 H100-80GB GPUs). Training time ranged from a few hours for the 7B model to up to 2.5 days for the final 32B model experiments.

\section{Limitations}
\label{sec:limitations}
It is important for each study—especially in the LLM domain—to clearly state its limitations. In our work, as noted in~\sref{sec:learning_from_trianing_stratety}, we did not exhaustively explore PPO hyperparameters, which may limit the strength of our conclusions. Similarly, we only partially examined data augmentation strategies~\sref{sec:learning_from_trianing_data}. In addition, while general-purpose LLMs are known to be prompt-sensitive, we did not systematically explore prompt variations~\sref{sec:learning_from_benchmark}. Finally, we have yet to evaluate \OURS across different model families to assess its generalizability.

\end{document}